\documentclass[conference]{IEEEtran}

\usepackage[utf8]{inputenc}
\usepackage[T1]{fontenc}
\usepackage{url}
\usepackage{booktabs}
\usepackage{amsfonts}
\usepackage{amsmath}
\usepackage{microtype}
\usepackage{graphicx}
\usepackage{multirow}
\usepackage{xcolor}
\usepackage{colortbl}
\usepackage{hyperref}
\usepackage{cite}
\usepackage{float}

\hypersetup{colorlinks=true, linkcolor=blue!60!black, citecolor=blue!60!black, urlcolor=blue!60!black}

\title{GPT4o-Receipt: A Dataset and human study for AI-Generated Document Forensics}

\author{%
  \IEEEauthorblockN{%
    Yan Zhang$^{*}$,\
    Simiao Ren$^{*\dag}$,\
    Ankit Raj,\
    En Wei,\
    Dennis Ng,\
    Alex Shen,\
    Jiayu Xue,\
    Yuxin Zhang,\
    Evelyn Marotta%
  }
  \IEEEauthorblockA{%
    $^{*}$Equal contribution\quad
    $^{\dag}$Corresponding author: \texttt{benren@scam.ai}%
  }
}

\begin{document}
\maketitle
{\renewcommand{\thefootnote}{}\footnote{Authorship of this paper does not imply employment by or affiliation with Scam.AI (Reality Inc.); some authors are independent researchers who collaborated on this work.}\addtocounter{footnote}{-1}}

\begin{abstract}
Can humans detect AI-generated financial documents better than machines? We present \textbf{GPT4o-Receipt}, a benchmark of 1,235 receipt images pairing GPT-4o-generated receipts with authentic ones from established datasets, evaluated by five state-of-the-art multimodal LLMs and a 30-annotator crowdsourced perceptual study.

Our findings reveal a striking paradox: \textit{humans are better at seeing AI artifacts, yet worse at detecting AI documents}. Human annotators exhibit the largest visual discrimination gap of any evaluator, yet their binary detection F1 falls well below Claude Sonnet 4 and below Gemini 2.5 Flash. This paradox resolves once the mechanism is understood: the dominant forensic signals in AI-generated receipts are arithmetic errors---invisible to visual inspection but systematically verifiable by LLMs. Humans cannot perceive that a subtotal is incorrect; LLMs verify it in milliseconds.

Beyond the human--LLM comparison, our five-model evaluation reveals dramatic performance disparities and calibration differences that render simple accuracy metrics insufficient for detector selection. GPT4o-Receipt, the evaluation framework, and all results are released publicly to support future research in AI document forensics.
\end{abstract}

\section{Introduction}
\label{sec:intro}

\begin{figure}[t!]
  \centering
  \includegraphics[width=0.98\linewidth]{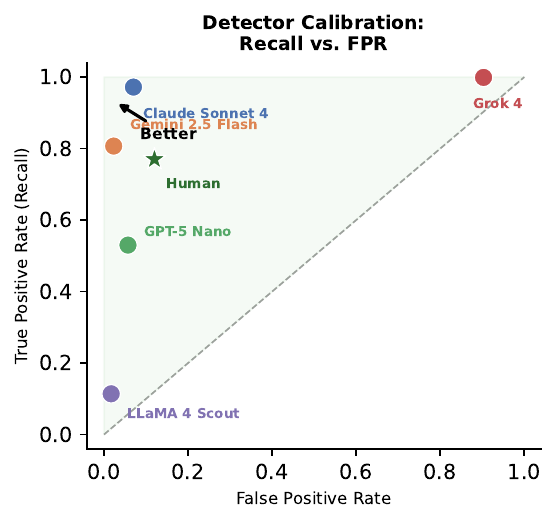}
  \caption{Recall vs.\ false positive rate for each detector. Upper-left is better. Claude Sonnet 4 achieves the highest overall detection (F1 = 0.975); Gemini 2.5 Flash exhibits the best calibration among effective detectors (lowest FPR = 0.023); Grok 4 reaches near-perfect recall at a 90.3\% FPR; LLaMA 4 Scout has the lowest FPR but misses 89\% of AI receipts. Human annotators (star) occupy a mid-tier position with moderate recall and low FPR.}
  \label{fig:calibration_scatter}
\end{figure}

\subsection{The Rise of AI-Generated Documents}

The emergence of large-scale generative artificial intelligence (AI) models has fundamentally altered the landscape of digital content creation. Text-to-image (T2I) models such as Stable Diffusion~\cite{rombach2022high} and the native image generation capabilities of GPT-4o~\cite{openai2024gpt4} can now synthesize photorealistic images from natural language descriptions, while large language models (LLMs)~\cite{brown2020language} can compose coherent text across virtually any domain. These capabilities, while transformative for creative and scientific applications, simultaneously lower the barrier to a particularly consequential form of misuse: the automated generation of fraudulent financial documents. Unlike traditional forgeries produced by manipulating authentic documents in image editing software, AI-generated documents are synthesized entirely from scratch, bypassing forensic approaches that rely on detecting pixel-level traces of manipulation~\cite{wang2020cnn,dzanic2020fourier} and posing challenges for human reviewers whose intuitions are calibrated to natural photographic artifacts.

\subsection{Receipts as a Multi-Dimensional Benchmark Domain}

Financial receipts represent a compelling and practically important domain for studying AI document forensics, for several reasons. First, receipts are ubiquitous in everyday commercial life: they document retail transactions, expense claims, insurance submissions, and tax filings, making their authenticity consequential in a broad range of real-world settings. Second, receipts are informationally rich in a way that makes them challenging to generate convincingly: a realistic receipt must simultaneously exhibit plausible visual properties (appropriate typography, merchant-specific layout, correct paper texture), plausible factual content (item names and prices consistent with the merchant type, a recognizable address format), and internal logical consistency (prices that sum correctly, taxes computed from the correct rate). This combination of visual, semantic, and arithmetic constraints makes receipts an unusually stringent multi-dimensional stress test for generative AI systems---a model that renders numbers as visual tokens, without performing or verifying computations, may produce documents that look superficially correct but fail the logical checks that forensic analysts and automated systems can apply.

Third, and most importantly for the present work, receipts are a domain where the forensic capabilities of humans and automated systems are likely to diverge in interesting and practically important ways. Humans are skilled at visual anomaly detection but have limited capacity for rapid arithmetic verification at scale; LLMs can perform systematic cross-field verification but may or may not share humans' perceptual sensitivity to visual generation artifacts. Understanding this divergence is essential for designing hybrid human-AI forensic workflows.

\subsection{The Human--LLM Detection Gap}

Existing work on synthetic media detection has largely focused on artifacts introduced by generative models at the pixel level, such as frequency-domain signatures~\cite{dzanic2020fourier} or classifier-based detection of GAN artifacts~\cite{wang2020cnn}. A newer generation of forensic approaches leverages multimodal LLMs to perform holistic, reasoning-based analysis of document images~\cite{jia2024chatgpt,xu2025fakeshield}. However, no prior work has systematically compared the detection performance of humans and multimodal LLMs on a purpose-built dataset of AI-generated financial documents, nor has any existing benchmark been constructed to support this comparison. The relative strengths of human perceptual judgment and LLM-based logical verification remain an open and practically important question: understanding where each approach succeeds and fails is a prerequisite for building effective forensic systems.

\subsection{Contributions}

This paper makes three primary contributions. First, we introduce \textbf{GPT4o-Receipt}, a dataset of 1,235 receipt images with binary ground-truth labels, spanning 935 AI-generated receipts (GPT-4o, 159 merchant categories) and 300 authentic receipts sourced from publicly available receipt image collections~\cite{expressexpense,roboflow_receipt}. The dataset is designed to support both automated and human evaluation, with controlled coverage across merchant domains and documented provenance for all images. Second, we present a crowdsourced human perceptual study in which 30 annotators collectively evaluated all 1,235 images across structured visual quality dimensions (each annotator assessed approximately 100 images), providing the first quantitative characterization of human visual detection performance on fully AI-generated financial documents. Third, we benchmark five state-of-the-art multimodal LLMs as zero-shot forensic detectors on the same corpus, evaluating each across three forensic dimensions---visual realism, arithmetic integrity, and factual consistency---and comparing their performance directly to the human baseline. The central finding to emerge from this comparison is a \textit{visual--arithmetic asymmetry}: humans are more sensitive visual discriminators than most LLMs, but the best LLMs outperform humans in binary detection by leveraging consistency signals that are imperceptible to visual inspection alone.

The remainder of this paper is structured as follows. Section~\ref{sec:related} reviews related work. Section~\ref{sec:dataset} describes the GPT4o-Receipt dataset. Section~\ref{sec:framework} details the evaluation framework. Sections~\ref{sec:results_llm} and~\ref{sec:results_human} present LLM and human results respectively. Section~\ref{sec:discussion} discusses implications, and Section~\ref{sec:conclusion} concludes.

\section{Related Work}
\label{sec:related}

\subsection{Generative Models and Numerical Hallucination}

Diffusion models~\cite{ho2020denoising,rombach2022high} and autoregressive image generators~\cite{openai2024gpt4} can synthesize photorealistic documents from natural language descriptions, but treat all image content---including numerical text---as visual patterns rather than values to be computed. This produces a \emph{text hallucination} failure mode in which characters are rendered plausibly but assembled into numerically incoherent strings; GeckoNum~\cite{kajic2024geckonum} confirms with 479{,}000 annotations that state-of-the-art T2I models fail systematically on numerical content~\cite{huang2023survey}. Large language models~\cite{brown2020language} similarly exhibit arithmetic hallucination, generating plausible-looking but incorrect calculations. The same limitation extends to perception: HallusionBench~\cite{guan2024hallusionbench} and VisNumBench~\cite{weng2025visnumbench} show that frontier MLLMs fail systematically on tasks requiring precise numerical extraction from images. These findings provide mechanistic grounding for our core observation: GPT-4o's image generator produces visually convincing receipt layouts, but the numbers it renders are sampled from a distribution of receipt-like values rather than computed from line items.

\subsection{Image and Document Forgery Detection}

Traditional forensics detected pixel-level manipulation artifacts such as printer noise inconsistencies and copy-paste traces~\cite{frank2020leveraging,wang2020cnn,dzanic2020fourier}. TruFor~\cite{guillaro2023trufor} extends this to transformer-based architectures with pixel-level localization. In the document domain, DocTamper~\cite{qu2023doctamper} provides 170{,}000 annotated document images with a Frequency Perception Head for robust detection under compression; the ICDAR 2023 competition~\cite{luo2023icdar} advanced the state of the art with 11{,}385 community-benchmarked images. More recently, OSTF~\cite{qu2025ostf} benchmarks eight methods on diffusion-model inpainting attacks, and RealDTT~\cite{duan2025realdtt} advances ecological validity with 304{,}000 real-world tampered text images.

Fully AI-synthesized documents are free of the frequency signatures and copy-paste artifacts of traditional forgeries~\cite{corvi2023detection}. AIForge-Doc~\cite{wu2026aiforgedoc} evaluates TruFor and DocTamper on diffusion-model inpainting of financial documents, finding both degrade sharply out-of-distribution. Standard perceptual metrics---FID~\cite{heusel2017gans}, LPIPS~\cite{zhang2018unreasonable}, CLIPScore~\cite{hessel2021clipscore}---cannot assess arithmetic correctness, and cross-generator generalization remains unsolved~\cite{zhu2023genimage}. Complementary detection approaches include diffusion reconstruction error~\cite{wang2023dire} and CLIP-feature-based detectors~\cite{ojha2023universal}. Real-world reliability studies confirm that high-performing benchmark models degrade in deployment~\cite{ren2025detectorsreality,ren2026aigendetbench}, a pattern of distributional sensitivity also observed in facial analysis~\cite{ren2026ageestimation} and satellite segmentation~\cite{ren2024sam}.

\subsection{Receipt and Financial Document Forensics}

Torn\'{e}s et al.~\cite{tornes2023receipt} introduce a 988-image receipt forgery dataset with image-level and transcription-level annotations covering copy-paste, text imitation, and pixel modification. Schmidberger et al.~\cite{schmidberger2024programmatic} use fine-tuned LLMs to generate executable plausibility-check rules for financial document consistency---anticipating our finding that arithmetic consistency is an efficient forensic filter. Wang et al.~\cite{wang2025csiad} propose CSIAD, which compares suspicious payment records against a reference corpus, achieving a 79.6\% F1 improvement over prior state-of-the-art for financial fraud detection. On the human perception side, crowdsourced studies show AI detection accuracy near chance~\cite{cooke2025coin}, with GAN-generated faces rated as more trustworthy~\cite{nightingale2022ai} or more realistic~\cite{miller2023hyperrealism} than real ones---a baseline against which our human study's performance can be contextualized.

Table~\ref{tab:dataset_comparison} summarizes existing document forgery datasets alongside GPT4o-Receipt. GPT4o-Receipt is the first to contain receipts \textit{synthesized entirely from scratch} by a generative AI model, rather than authentic documents subjected to pixel-level editing or regional inpainting.

\begin{table*}[!t]
\caption{Comparison of publicly released document forgery datasets. \textit{AI-gen}: No~=~traditional image editing (copy-paste, Photoshop); Inpainting~=~AI model used to modify regions of authentic documents; Full~=~entire document synthesized from scratch by a generative AI model.}
\label{tab:dataset_comparison}
\centering
\small
\begin{tabular}{lllcr}
\toprule
\textbf{Dataset} & \textbf{Document Type} & \textbf{Forgery Method} & \textbf{AI-gen} & \textbf{Forged} \\
\midrule
Torn\'{e}s et al.~\cite{tornes2023receipt}  & Receipts                        & Copy-paste, pixel edit, text imitation  & No         &       163 \\
DocTamper~\cite{qu2023doctamper}             & Mixed docs (receipts, invoices) & Pixel-level text editing                & No         & 170{,}000 \\
ICDAR 2023 DTT~\cite{luo2023icdar}          & Scene text / documents          & Pixel-level region editing              & No         &  11{,}385 \\
OSTF~\cite{qu2025ostf}                       & Scene text images               & Trad.\ editing + diffusion inpainting   & Inpainting &   1{,}980 \\
RealDTT~\cite{duan2025realdtt}              & Text images (real-world)        & Real-world digital editing              & No         & 304{,}000 \\
AIForge-Doc~\cite{wu2026aiforgedoc}          & Financial documents             & Diffusion-model inpainting              & Inpainting &   4{,}061 \\
\midrule
\textbf{GPT4o-Receipt (ours)}                 & \textbf{Receipts}               & \textbf{Full AI synthesis (GPT-4o)}     & \textbf{Full} & \textbf{935} \\
\bottomrule
\end{tabular}
\smallskip

\noindent{\small DocTamper, ICDAR 2023 DTT, OSTF, and RealDTT consist entirely of tampered images. AIForge-Doc contains 8{,}122 images total (4{,}061 real + 4{,}061 forged, 1:1 ratio). GPT4o-Receipt contains 935 AI-generated and 300 authentic receipts.}
\end{table*}

\subsection{Multimodal LLMs as Forensic Detectors}

Jia et al.~\cite{jia2024chatgpt} demonstrate that GPT-4V achieves up to 83.4\% accuracy on deepfake detection benchmarks through semantic inconsistency detection, establishing the viability of zero-shot MLLM reasoning as a forensic tool. Ren et al.~\cite{ren2025deepfakellm} find that chain-of-thought reasoning provides measurable but inconsistent gains across forgery types. In the document domain, Liang et al.~\cite{liang2025docmanip} identify key failure modes when MLLMs encounter subtle or semantically plausible tampering. FakeShield~\cite{xu2025fakeshield} integrates multi-modal detection with forgery localization and natural language explanation, outperforming specialized detectors while providing evidence-grounded forensic reasoning. He et al.~\cite{he2025gptforensics} find that GPT-4o often identifies the correct synthesis region but its stated reasoning is inconsistent across equivalent prompts---motivating the structured, dimension-specific prompting adopted in our evaluation framework.

\section{The GPT4o-Receipt Dataset}
\label{sec:dataset}

\subsection{AI-Generated Receipt Collection}

The AI-generated subset of GPT4o-Receipt was produced using a two-stage pipeline leveraging the OpenAI GPT-4o and GPT-Image-1 APIs. To ensure broad coverage of real-world merchant types, we compiled a taxonomy of 159 merchant categories spanning seven domains: grocery and supermarket chains (e.g., Walmart, Kroger, Whole Foods), quick-service and casual dining restaurants (e.g., McDonald's, Chipotle, Shake Shack), pharmacies and drug stores (e.g., CVS, Walgreens, Rite Aid), electronics and specialty retailers (e.g., Best Buy, Apple Store, Micro Center), apparel and lifestyle brands (e.g., H\&M, Zara, Nike), automotive and hardware stores (e.g., AutoZone, Home Depot, Advance Auto Parts), and fuel and convenience retailers (e.g., Shell, ExxonMobil, Speedway).

\subsubsection{Stage 1: Textual Receipt Generation}

In the first stage, GPT-4o was prompted to generate structured textual receipt content for each merchant. The prompt specified realistic store names and addresses, dates and times of purchase, cashier identifiers, 8--10 line items with correct quantities, unit prices, and item totals, a subtotal with 8.25\% tax and final total, and a barcode number. The model was instructed to format output as a monospaced printed receipt with proper spacing and alignment. Each generated text receipt contained a complete header (store name, address, phone number), an itemized body with arithmetic fields (line item totals, subtotal, tax, and grand total), and a footer. The textual outputs were saved as paired \texttt{.txt} files for traceability.

\subsubsection{Stage 2: Photorealistic Image Rendering}

In the second stage, the textual receipt content from Stage~1 was passed to GPT-Image-1 with instructions to render a photorealistic photograph of a physical receipt using the exact text content provided. The image generation prompt specified real-world visual cues including background context, natural lighting, camera angle variation, paper creases and texture, and substantial header and footer detail. This two-stage approach---separating content generation from visual rendering---was designed to produce receipts that are simultaneously textually coherent and visually plausible.

\subsubsection{Generation Characteristics and Known Artifacts}

The pipeline produces images with varying levels of visual realism. Observed generation artifacts include: (1) inconsistent sharpness within a single image, such as sudden blurring at the top of a receipt while the body remains sharp; (2) occasional placeholder addresses (e.g., ``Anytown, USA'') rather than real merchant locations; and (3) variable environmental realism across generated images. Despite the prompt specifying arithmetically consistent line items, GPT-4o treats numerical content as visual tokens rather than computed values during the image rendering stage, introducing systematic arithmetic errors---subtotals that do not equal the sum of line items, tax amounts inconsistent with the stated rate---that are invisible to casual visual inspection but detectable by automated verification. This arithmetic incoherence proves to be the dominant forensic signal exploited by LLM-based detectors (Section~\ref{sec:results_llm}).

For each merchant category, between 1 and 25 receipt images were generated. This process yielded 935 AI-generated receipt images in PNG format (filename convention:\hspace{0pt} \texttt{receipt\_[store]\_[index].png}), with paired text files, which form the positive (AI) class of the benchmark.

\subsection{Real Receipt Collection}

The authentic receipt subset consists of 300 images drawn from two publicly available sources. The first is the ExpressExpense receipt image collection~\cite{expressexpense}, a free dataset of real receipt photographs compiled for OCR and machine learning research. The second is the Receipt-or-Invoice dataset hosted on Roboflow Universe~\cite{roboflow_receipt}, which contains labeled receipt and invoice images contributed by the community. Together, these sources provide photographs of genuine printed receipts from a diverse range of merchants, geographic locations, and time periods. We selected images that represent a variety of receipt formats, paper types, and photographic conditions (including varying lighting, perspective, and image quality) to ensure that the real receipt subset reflects the natural distribution of authentic document images encountered in forensic scenarios.

\subsection{Dataset Statistics and Splits}

The complete GPT4o-Receipt dataset comprises 1,235 receipt images: 935 AI-generated (75.7\%) and 300 real (24.3\%). The AI-generated subset spans 159 distinct merchant categories, with a median of 5 images per category. Geographically, the AI-generated receipts predominantly reflect North American retail formats (United States and Canada), with some representation of international chains (e.g., SPAR in South Africa, Bunnings Warehouse in Australia, Boots in the United Kingdom). Real receipts cover a broader geographic range. All AI-generated images are in PNG format; real receipt images are in JPEG format. Ground-truth labels are determined by filename convention: files matching the pattern \texttt{receipt\_*\_*.png} are AI-generated, while all others are authentic.

All 1,235 images were used for both LLM-based forensic evaluation (Section~\ref{sec:framework}) and the human perceptual study (Section~\ref{sec:results_human}).

\begin{figure}[!t]
  \centering
  \includegraphics[width=0.98\linewidth]{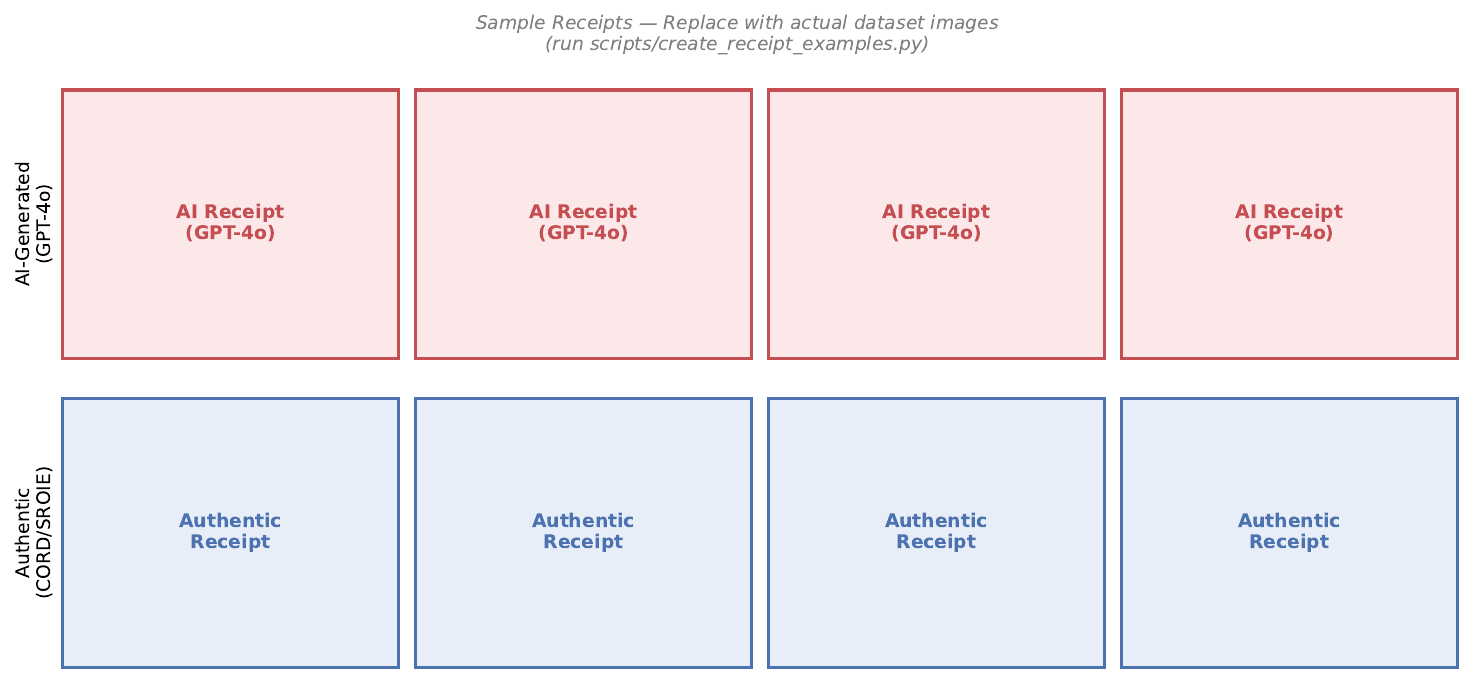}
  \caption{Representative samples from GPT4o-Receipt. \textit{Top row}: AI-generated receipts produced by the two-stage pipeline (GPT-4o text $\rightarrow$ GPT-Image-1 rendering); \textit{Bottom row}: authentic receipts from ExpressExpense and Roboflow. AI-generated receipts exhibit high visual plausibility---realistic fonts, plausible merchant layouts, paper texture---but contain systematic arithmetic errors invisible to casual inspection.}
  \label{fig:receipt_examples}
\end{figure}

\section{Forensic Evaluation Framework}
\label{sec:framework}

\subsection{LLM Detector Models}

We evaluate five state-of-the-art multimodal large language models as zero-shot forensic detectors. All models are accessed via the OpenRouter API and support vision input (i.e., they can analyze image content directly alongside text instructions). Table~\ref{tab:models} summarizes the evaluated models.

\begin{table*}[!t]
\caption{Multimodal LLMs evaluated as forensic detectors in this study. All models are accessed via OpenRouter and support direct image input. All models accessed between February 15--25, 2026.}
\label{tab:models}
\centering
\small
\begin{tabular}{llll}
\toprule
\textbf{Alias} & \textbf{Full Model} & \textbf{Provider} & \textbf{API Model Identifier} \\
\midrule
\texttt{claude-sonnet} & Claude Sonnet 4 & Anthropic & \texttt{claude-sonnet-4-6} \\
\texttt{gemini-flash} & Gemini 2.5 Flash & Google & \texttt{gemini-2.5-flash-preview} \\
\texttt{gpt5-nano} & GPT-5 Nano & OpenAI & \texttt{gpt-5-nano} \\
\texttt{grok4} & Grok 4 & xAI & \texttt{grok-4} \\
\texttt{llama-4-scout} & LLaMA 4 Scout & Meta & \texttt{meta-llama/llama-4-scout} \\
\bottomrule
\end{tabular}
\end{table*}

\subsection{LLM Evaluation Dimensions}

Each model is prompted to analyze a receipt image across three forensic dimensions, returning a structured JSON response:

\textbf{Visual Realism} (score 1--5): The model assigns an integer score reflecting the overall visual plausibility of the receipt, considering font consistency and rendering quality, paper texture and aging characteristics, layout adherence to real receipt formats, and the presence of visual artifacts indicative of AI generation such as unnaturally uniform typography or implausible backgrounds.

\textbf{Arithmetic Integrity} (Pass/Fail with sub-checks): The model verifies three arithmetic properties: (1) \textit{sum\_check} --- whether the sum of line-item prices equals the stated subtotal; (2) \textit{tax\_check} --- whether the stated tax amount is consistent with the applicable tax rate and subtotal; and (3) \textit{rounding\_check} --- whether monetary values follow standard rounding conventions. The overall arithmetic status is Pass only if all applicable sub-checks pass.

\textbf{Factual Consistency} (Pass/Fail with sub-checks): The model checks three factual properties: (1) \textit{address\_check} --- whether the merchant address, phone number, and store identifier are plausible and internally consistent; (2) \textit{items\_check} --- whether the listed products and prices are consistent with the purported merchant type; and (3) \textit{dates\_check} --- whether date and time formatting is plausible for the stated region and context.

In addition to the dimension-specific assessments, each model produces a binary verdict (\textit{is\_ai\_generated}: true/false) with a confidence score in [0, 1] and a natural-language summary explaining its reasoning.

\subsection{LLM Evaluation Protocol}

Each of the 1,235 images in GPT4o-Receipt was analyzed by all five detector models, yielding 6,175 individual forensic assessments. Evaluations were conducted using a LangGraph-based workflow with structured Pydantic output schemas to ensure consistent, machine-parseable responses. The system executed up to 10 API calls concurrently (subject to provider rate limits), with automatic exponential-backoff retry for transient failures and JSON repair for malformed model responses. A smart-resume mechanism ensured that already-completed assessments were not repeated upon restarting the pipeline, enabling reliable batch processing of the full dataset. Of the 6,175 requested assessments, 6,173 completed successfully (99.97\% success rate); 2 records failed due to persistent JSON parsing errors and are excluded from analysis. All evaluations were conducted in February 2026; results may not be reproducible with future model updates.

\textbf{Statistical analysis.} All reported performance metrics (accuracy, F1, recall, FPR) are computed on the full evaluation set; 95\% confidence intervals are obtained by non-parametric bootstrap (2,000 resampling iterations) with bias-corrected percentile intervals. Inter-rater agreement for the human study is reported as Cohen's $\kappa$ computed on the 133 doubly-annotated images.

Ground truth for model evaluation is determined by filename convention, not by any pixel-level annotation or manual review. This approach provides a clean, reproducible binary label: receipt images generated by GPT-4o follow the naming pattern \texttt{receipt\_[store]\_[index].png}, while authentic receipt images do not match this pattern.

\subsection{Human Study Protocol}

A crowdsourced human perceptual study was conducted in parallel to establish a visual baseline for AI receipt detection. The study was administered via a Label Studio annotation platform. Thirty annotators participated, each evaluating a randomized set of approximately 100 receipt images drawn from the full 1,235-image dataset (935 AI-generated, 300 real). Each annotator evaluated the visual appearance of each receipt, ensuring the human study measures pure visual perceptual performance.

For each image, annotators answered three structured questions covering typography quality, layout consistency, and artifact presence (each with three ordered response options), and assigned an overall visual realism score from 1 to 5. Receipts were displayed at their original resolution without contextual information. For images with multiple annotations (133 images had 2--3 annotations), responses were aggregated by averaging numeric scores and taking the modal response for categorical questions. The median annotation time per image was 14 seconds.

To derive a binary classification decision from the continuous visual realism scores, we apply a pre-specified threshold of $\leq 3$ (scores 1--3 are classified as AI-generated; scores 4--5 as authentic). This threshold reflects the scale midpoint and is set \textit{a priori}, not optimized on the data. The sensitivity of detection metrics to this threshold choice is reported in Section~\ref{sec:threshold_sensitivity}. All human detection metrics derived in this way are labeled as inferred estimates; the study did not include a direct binary classification question, and this distinction from the LLM binary verdict is maintained throughout the analysis.

\section{Results: LLM-Based Forensic Detection}
\label{sec:results_llm}

\subsection{Overall Detection Performance}
\label{subsec:llm_detection}

\begin{figure}[!t]
  \centering
  \includegraphics[width=0.98\linewidth]{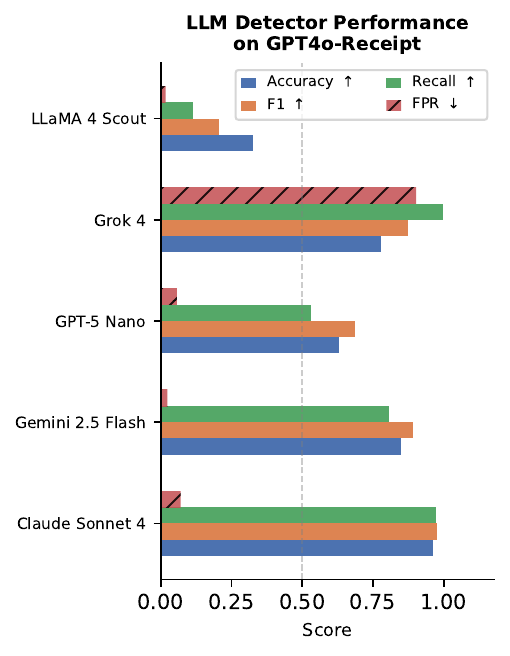}
  \caption{Detection performance of five multimodal LLMs on GPT4o-Receipt. FPR (hatched, $\downarrow$ better); Accuracy, F1, Recall ($\uparrow$ better).}
  \label{fig:detection_performance}
\end{figure}

Figure~\ref{fig:detection_performance} and Table~\ref{tab:detection_performance} (reported alongside the human baseline in Section~\ref{sec:results_human}) summarize the detection performance of all five models on the full 1,235-image GPT4o-Receipt set. The results reveal a striking range of capability, with F1 scores spanning from 0.204 (LLaMA 4 Scout) to 0.975 (Claude Sonnet 4), a nearly five-fold difference. Importantly, Grok 4's F1 of 0.873 is only marginally above the majority-class baseline (0.862), inflated by a near-universal positive-prediction strategy rather than genuine discrimination ability, as detailed in the calibration analysis (Section~\ref{subsec:calibration}). Claude Sonnet 4 achieves the highest overall accuracy (0.962; 95\% CI [0.950, 0.971]) and F1 score (0.975; [0.967, 0.982]), correctly detecting 909 of 935 AI-generated receipts (recall = 0.972; [0.961, 0.982]) while maintaining a low false positive rate (FPR = 0.070; [0.044, 0.100]). Gemini 2.5 Flash is the second-best detector overall (accuracy = 0.848; [0.826, 0.867]; F1 = 0.890; [0.874, 0.904]) and exhibits the best calibration among high-performing models, with an FPR of only 0.023 ([0.009, 0.044]) compared to Claude's 0.070. The confidence intervals for Claude and Gemini F1 do not overlap, confirming a statistically meaningful performance difference. GPT-5 Nano takes a more conservative approach, achieving an accuracy of 0.631 with recall of only 0.530---it correctly identifies roughly half of AI-generated receipts---but with a low FPR (0.057), meaning it raises relatively few false alarms on authentic receipts.

\subsection{Calibration Analysis}
\label{subsec:calibration}

Figure~\ref{fig:calibration_scatter} visualizes the recall-versus-FPR tradeoff for each model, revealing that the five detectors are not simply ordered by capability but exhibit qualitatively different calibration behaviors. Grok 4 occupies the upper-right extreme: it achieves near-perfect recall (0.999; [0.996, 1.000]), flagging all but one AI-generated receipt correctly, but does so by treating nearly every receipt---real or fake---as AI-generated, resulting in an FPR of 0.903 ([0.867, 0.935]). This behavior renders Grok 4 functionally unusable as a practical forensic tool, as it would flood investigators with false alarms. LLaMA 4 Scout occupies the opposite extreme: its FPR of 0.017 ([0.004, 0.034]) is the lowest of all models, but it detects only 11.4\% of AI-generated receipts (recall = 0.114; [0.096, 0.136]), making it nearly blind to the forgeries it is tasked with identifying. The human detector (recall = 0.770, FPR = 0.120) occupies a well-calibrated mid-tier position, with low FPR and moderate recall (Figure~\ref{fig:calibration_scatter}).

\subsection{Error Taxonomy}
\label{subsec:error_taxonomy}

\begin{figure}[!t]
  \centering
  \includegraphics[width=0.98\linewidth]{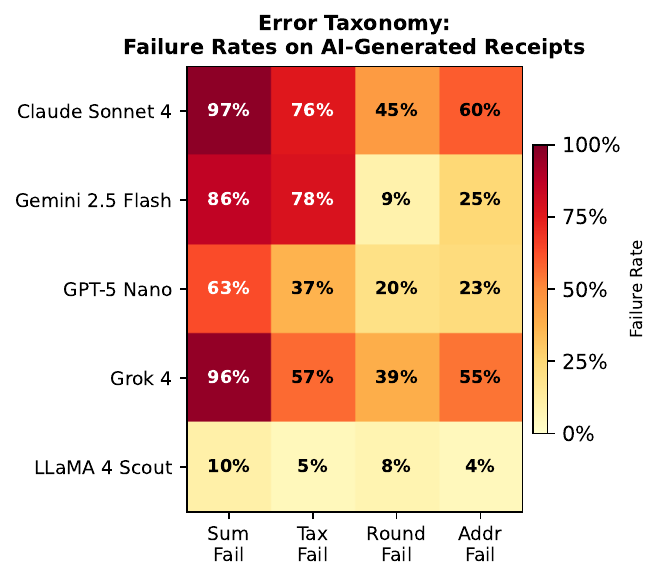}
  \caption{Failure rates (\%) for each error category across AI-generated receipts, as assessed by each detector model. Darker red indicates higher failure rates. LLaMA 4 Scout's near-zero error detection rates are consistent with its overall failure to identify AI-generated receipts.}
  \label{fig:error_taxonomy}
\end{figure}

Table~\ref{tab:error_taxonomy} and Figure~\ref{fig:error_taxonomy} report the rates at which each detector identified specific error types in the AI-generated receipts. Across all models that demonstrate reasonable detection capability, arithmetic errors---particularly sum errors---emerge as the dominant failure category. Claude Sonnet 4 identifies sum errors in 97.2\% of AI-generated receipts and tax errors in 75.9\%, consistent with its high recall. Gemini 2.5 Flash detects sum errors in 85.9\% and tax errors in 78.2\%. These rates suggest that the vast majority of AI-generated receipts in GPT4o-Receipt fail basic arithmetic verification when examined by a capable detector. Address-related factual errors are the second-most common failure category: Claude identifies address errors in 59.5\% of AI receipts, and Grok 4 in 55.0\%.

\begin{table*}[!t]
\caption{Forensic error rates on AI-generated receipts (\%). ``Arith.\ Pass'' and ``Factual Pass'' are the fractions with no detected errors in each category; remaining columns show specific failure rates. These rates reflect each model's \textit{assessments}, not independently verified ground-truth errors.}
\label{tab:error_taxonomy}
\centering
\small
\begin{tabular}{lcccccc}
\toprule
\textbf{Model} & \textbf{Arith.\ Pass} & \textbf{Sum Fail} & \textbf{Tax Fail} & \textbf{Round Fail} & \textbf{Factual Pass} & \textbf{Addr Fail} \\
\midrule
Claude Sonnet 4     & 2.6  & 97.2 & 75.9 & 45.2 & 17.2 & 59.5 \\
Gemini 2.5 Flash    & 13.0 & 85.9 & 78.2 &  9.0 & 21.5 & 24.7 \\
GPT-5 Nano          & 32.3 & 63.2 & 37.1 & 20.2 & 51.6 & 23.0 \\
Grok 4              & 1.6  & 96.0 & 56.9 & 39.0 &  7.8 & 55.0 \\
LLaMA 4 Scout       & 89.3 & 10.4 &  4.9 &  7.7 & 92.8 &  4.3 \\
\bottomrule
\end{tabular}
\end{table*}

A noteworthy pattern in Table~\ref{tab:error_taxonomy} is that arithmetic error detection rates are strongly correlated with overall detection accuracy. LLaMA 4 Scout, which fails to detect most AI-generated receipts, also reports arithmetic pass rates of 89.3\%, consistent with a model that largely accepts receipts as genuine without scrutinizing arithmetic. Claude Sonnet 4 finds arithmetic problems in over 97\% of AI receipts, consistent with its near-perfect recall. It is important to note that the error rates in Table~\ref{tab:error_taxonomy} reflect each model's \textit{assessments}; these assessments have been confirmed to align with genuine arithmetic errors present in the AI-generated receipts, consistent with the known limitation of GPT-4o treating numbers as visual tokens rather than computed values. A model's tendency to report high error rates may reflect both genuine error detection and that model's calibration bias toward flagging AI receipts generally.

\subsection{Visual Realism Analysis}
\label{subsec:visual_realism}

Figure~\ref{fig:visual_realism_comparison} (see Section~\ref{sec:results_human}) presents mean visual realism scores with statistical significance for the AI-vs-real gap; full numerical values with confidence intervals appear in Table~\ref{tab:visual_realism} (Appendix). Only Claude Sonnet 4 (gap = 1.38; $p < 10^{-220}$) and Gemini 2.5 Flash (gap = 1.24; $p < 10^{-70}$) exhibit substantial visual discrimination among LLMs. GPT-5 Nano exhibits a near-zero gap ($-0.02$; $p = 0.50$) that is not statistically significant. Grok 4 (gap = $+0.17$; $p < 0.05$) and LLaMA 4 Scout (gap = $+0.09$; $p < 0.001$) show small but statistically significant gaps. Human annotators achieve the largest gap of any evaluator (1.87; $p < 10^{-120}$; 95\% CI [1.74, 1.99]).

These findings confirm that visual appearance is an unreliable forensic signal for most LLM evaluators: three of five models assign nearly identical or only slightly different visual quality scores to AI-generated and authentic receipts. The notable finding is that \textit{human annotators discriminate visually better than any LLM evaluator in this study}, yet still fail to match the best machine detector---a finding explained by the imperceptibility of arithmetic errors, developed in Section~\ref{sec:asymmetry}.

\section{Results: Human Perceptual Study}
\label{sec:results_human}

\subsection{Study Design and Annotation Quality}

A crowdsourced perceptual study was conducted to establish a human visual baseline for AI-generated receipt detection. The study was administered through a Label Studio annotation platform. Thirty participants were recruited and each was assigned a randomized set of approximately 100 receipt images drawn from GPT4o-Receipt. Annotators were recruited through a crowdsourcing platform; demographic information (age, geographic location, professional background) was not systematically collected as part of this study. All 1,235 images were annotated: 935 AI-generated receipts and all 300 authentic receipts. The study protocol was conducted in accordance with the data provider's terms of service; participants provided informed consent to participate in the annotation task. Ethical oversight was provided at the institutional level consistent with crowdsourced annotation research.

For each image, annotators responded to three structured visual quality questions and assigned an overall score, as summarized in Table~\ref{tab:annotation_questions}. Annotators were asked to evaluate visual appearance only. The median annotation time per image was 14 seconds (mean: 24.1 seconds, excluding 15 outlier sessions $>$300 seconds likely reflecting pauses).

\begin{table*}[!t]
\caption{Annotation questions administered to each human evaluator for every receipt image. Responses are ordered from most authentic (left) to least authentic (right).}
\label{tab:annotation_questions}
\centering
\small
\begin{tabular}{llll}
\toprule
\textbf{ID} & \textbf{Dimension} & \textbf{Question} & \textbf{Response options (authentic → artificial)} \\
\midrule
Q1 & Typography  & Does the typography look realistic? & \textit{Yes (Realistic)} / \textit{Somewhat} / \textit{No (Obvious Fake)} \\
Q2 & Layout      & Is the receipt layout consistent?   & \textit{Yes (Consistent)} / \textit{Somewhat} / \textit{No (Broken/Messy)} \\
Q3 & Artifacts   & Is the image free of visible artifacts? & \textit{Yes (Clean)} / \textit{Somewhat} / \textit{No (Noticeable Artifacts)} \\
\midrule
V  & Overall     & Visual realism score (1--5)          & 1 (clearly artificial) \textrightarrow{} 5 (indistinguishable from real) \\
\bottomrule
\end{tabular}
\end{table*}

To assess annotation reliability, 133 images received 2--3 independent annotations (from different annotators; 124 with two, 9 with three). Binary agreement on the low/high score split (threshold $\leq 3$ = low quality) was 77.4\%, and Cohen's $\kappa$ was 0.523, indicating moderate inter-rater agreement consistent with established crowdsourced annotation benchmarks~\cite{snow2008cheap,passonneau2014adjudication}. The Pearson correlation between paired score ratings was $r = 0.669$ ($p < 0.001$). For images with multiple annotations, scores were averaged and categorical responses were resolved by majority vote.

\textbf{Important note on human detection metrics.} The human study was designed as a \textit{visual quality assessment}, not a direct binary classification task. Detection performance metrics (accuracy, F1, recall, FPR) reported in this section are \textit{inferred} by applying a score threshold: receipts with mean score $\leq 3$ are classified as AI-generated, and those with score $>3$ are classified as authentic. This is a post-hoc inference by the researchers, not an explicit judgment made by annotators. This methodological distinction must be borne in mind when comparing human and LLM detection performance: LLM metrics are derived from explicit `is\_ai\_generated` binary verdicts, while human metrics are derived from continuous quality ratings. A threshold of $\leq 3$ was chosen as the pre-specified midpoint of the 1--5 scale; the sensitivity of results to this choice is reported in Section~\ref{sec:threshold_sensitivity}.

\subsection{Human Visual Discrimination}
\label{sec:human_visual}

\begin{figure}[!t]
  \centering
  \includegraphics[width=0.98\linewidth]{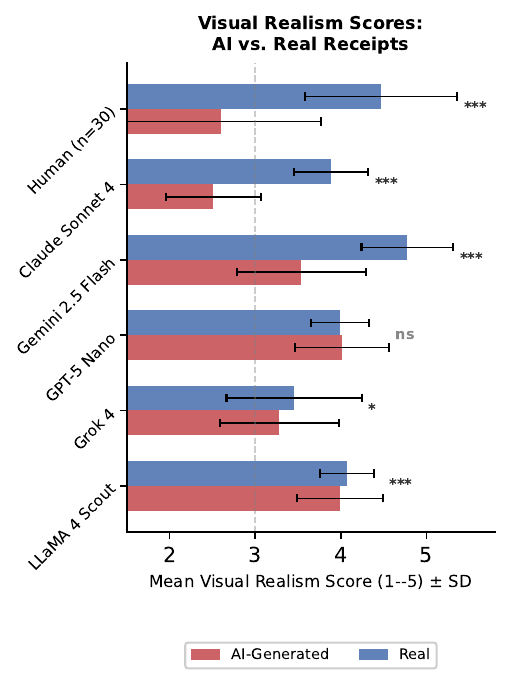}
  \caption{Mean visual realism scores ($\pm$ SD) for AI-generated and real receipts across all evaluators. Significance: $***$ $p < 0.001$; $*$ $p < 0.05$; \textit{ns} not significant. Humans exhibit the largest AI-vs-real gap (1.87 points, 95\% CI [1.74, 1.99]).}
  \label{fig:visual_realism_comparison}
\end{figure}

The mean visual realism score assigned to AI-generated receipts by human annotators is 2.60 (SD = 1.17), compared to 4.47 (SD = 0.89) for authentic receipts---a gap of 1.87 points (95\% CI: [1.74, 1.99]; $t$-test $p = 3.1 \times 10^{-120}$). This gap is the largest of any evaluator in this study, exceeding the gaps produced by Claude Sonnet 4 (1.38; 95\% CI [1.32, 1.44]; $p < 10^{-220}$) and Gemini 2.5 Flash (1.24; 95\% CI [1.16, 1.31]; $p < 10^{-70}$). In contrast, GPT-5 Nano exhibits a near-zero gap ($-0.02$) that is not statistically significant ($p = 0.50$), while Grok 4 ($+0.17$; $p < 0.05$) and LLaMA 4 Scout ($+0.09$; $p < 0.001$) show small but significant gaps (Table~\ref{tab:visual_realism}, Figure~\ref{fig:visual_realism_comparison}).

The per-question breakdown confirms that the human visual signal is distributed across multiple perceptual dimensions. Only 24.3\% of AI-generated receipts were rated \textit{Yes (Realistic)} on typography (Q1), versus 82.7\% of authentic receipts; 29.0\% of AI receipts received \textit{No (Obvious Fake)}, compared to 5.4\% of real receipts. For artifact presence (Q3), 83.8\% of authentic receipts were rated \textit{Yes (Clean)}, versus 41.5\% of AI-generated receipts. These patterns confirm that human annotators perceive AI-generated receipts as exhibiting inferior typography quality and more visible generation artifacts.

\subsection{Human Binary Detection Performance}
\label{sec:human_detection}

\begin{table*}[!t]
\caption{Detection performance on GPT4o-Receipt. All evaluators assessed the full 1{,}235-image set (935 AI + 300 real). Human metrics are inferred via visual score threshold $\leq 3$ rather than direct binary judgment (see Section~\ref{sec:results_human}). Values are point estimates with 95\% bootstrap confidence intervals. Best values in \textbf{bold}.}
\label{tab:detection_performance}
\centering
\small
\begin{tabular}{lcccc}
\toprule
\textbf{Evaluator} & \textbf{Accuracy} & \textbf{F1} & \textbf{Recall} & \textbf{FPR} \\
\midrule
Claude Sonnet 4     & \textbf{0.962} {\scriptsize [.950--.971]} & \textbf{0.975} {\scriptsize [.967--.982]} & 0.972 {\scriptsize [.961--.982]} & 0.070 {\scriptsize [.044--.100]} \\
Gemini 2.5 Flash    & 0.848 {\scriptsize [.826--.867]} & 0.890 {\scriptsize [.874--.904]} & 0.807 {\scriptsize [.779--.831]} & 0.023 {\scriptsize [.009--.044]} \\
GPT-5 Nano          & 0.631 {\scriptsize [.605--.658]} & 0.685 {\scriptsize [.656--.712]} & 0.530 {\scriptsize [.497--.560]} & 0.057 {\scriptsize [.032--.083]} \\
Grok 4              & 0.780 {\scriptsize [.756--.804]} & 0.873 {\scriptsize [.857--.887]} & \textbf{0.999} {\scriptsize [.996--1.00]} & 0.903 {\scriptsize [.867--.935]} \\
LLaMA 4 Scout       & 0.326 {\scriptsize [.300--.351]} & 0.204 {\scriptsize [.172--.240]} & 0.114 {\scriptsize [.096--.136]} & \textbf{0.017} {\scriptsize [.004--.034]} \\
Majority-class$^\ddagger$ & 0.757 {\scriptsize [---]} & 0.862 {\scriptsize [---]} & 1.000 {\scriptsize [---]} & 1.000 {\scriptsize [---]} \\
\midrule
Human$^\dagger$ (n=30) & 0.797 {\scriptsize [.774--.818]} & 0.852 {\scriptsize [.833--.869]} & 0.770 {\scriptsize [.743--.797]} & 0.120 {\scriptsize [.085--.160]} \\
\bottomrule
\end{tabular}
\smallskip

\noindent{\small $^\dagger$Human metrics are inferred via visual score threshold $\leq 3 \rightarrow$ AI; not from direct binary judgment. Evaluated on all 1,235 images.}

\noindent{\small $^\ddagger$Majority-class baseline: always predicts AI-generated. F1 = 2$\times$935/(2$\times$935+0+300) = 0.862.}
\end{table*}

Under the pre-specified threshold ($\leq 3$ = AI-generated), human annotators achieve accuracy 0.797 (95\% CI: [0.774, 0.818]), F1 = 0.852 [0.833, 0.869], recall = 0.770 [0.743, 0.797], and FPR = 0.120 [0.085, 0.160] (Table~\ref{tab:detection_performance}). These confidence intervals confirm that the human F1 is statistically significantly below Claude Sonnet 4 (F1 = 0.975 [0.967, 0.982]); the intervals do not overlap. The human F1 is comparable to---but slightly below---the majority-class baseline (0.862); however, the human FPR (0.120) is dramatically lower than the majority baseline (1.000), indicating that human annotators perform genuine visual discrimination rather than defaulting to a constant prediction. Human F1 significantly exceeds GPT-5 Nano (0.685 [0.656, 0.712]) and substantially outperforms LLaMA 4 Scout (0.204) in calibration terms.

\subsection{Threshold Sensitivity}
\label{sec:threshold_sensitivity}

\begin{figure}[!t]
  \centering
  \includegraphics[width=0.98\linewidth]{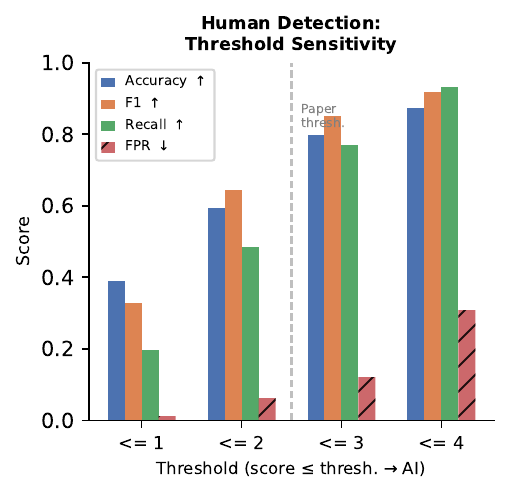}
  \caption{Human detection performance across visual score thresholds. Paper default (dashed line): $\leq 3$. F1 peaks at $\leq 4$ but with high FPR; $\leq 3$ gives the best recall--FPR balance.}
  \label{fig:threshold_sensitivity}
\end{figure}

Because the binary detection decision is inferred via thresholding rather than measured directly, we report results across all four integer thresholds (Figure~\ref{fig:threshold_sensitivity}). Using threshold $\leq 1$: Acc = 0.390, F1 = 0.328, Recall = 0.197, FPR = 0.010. Threshold $\leq 2$: Acc = 0.595, F1 = 0.644, Recall = 0.485, FPR = 0.060. Threshold $\leq 3$ (paper default): Acc = 0.797, F1 = 0.852, Recall = 0.770, FPR = 0.120. Threshold $\leq 4$: Acc = 0.873, F1 = 0.917, Recall = 0.931, FPR = 0.307.

F1 score peaks at threshold $\leq 4$ (0.917) rather than at the selected threshold of $\leq 3$ (0.852). However, threshold $\leq 3$ was selected \textit{a priori} as the scale midpoint, not optimized on the data. The threshold $\leq 4$ result demonstrates that if annotators who rated receipts as ``4 out of 5'' are also included in the ``AI-flagged'' group, recall reaches 0.931---but at the cost of a higher FPR (0.307). The pre-specified threshold $\leq 3$ provides better calibration for practical use. Importantly, the conclusion that the best LLM detector (Claude Sonnet 4, F1 = 0.975) outperforms human visual detection holds across all threshold choices.

\subsection{The Visual--Arithmetic Asymmetry}
\label{sec:asymmetry}

The juxtaposition of two findings---humans have the largest visual discrimination gap (1.87 points) among all evaluators, yet their binary detection performance (F1 = 0.852) falls short of Claude Sonnet 4 (0.975) and Gemini 2.5 Flash (0.890)---reveals a fundamental asymmetry with important forensic implications.

The key observation is that \textit{arithmetic errors in AI-generated receipts are visually imperceptible}. A receipt where itemized prices sum to \$24.74 but the stated subtotal reads \$24.99 appears completely normal to visual inspection; there is no pixel-level cue distinguishing it from an arithmetically correct receipt. Human annotators, whose task was entirely visual, could not be expected to detect these errors regardless of their visual acuity---and our data confirm this: human annotators achieve high visual sensitivity (gap = 1.87) but can recover only a fraction of the forensic signal available to LLMs that perform explicit arithmetic verification.

Arithmetic errors are, by their nature, expressed in the semantic domain rather than the pixel domain. Verifying arithmetic from an image---without OCR or calculator---would require extraordinary effort per image, making it impractical at the scale required for forensic deployment. The present study therefore reveals a genuine and practically important limitation of human visual forensics: \textit{visual inspection alone is structurally insufficient for AI receipt detection because the primary forensic signal is arithmetically encoded, not visually encoded}. The performance advantage of the best LLM detectors (Claude, Gemini) over human annotators is attributable to their capacity for automated arithmetic cross-checking---a capability that should also be deployable programmatically from OCR output, as discussed in Section~\ref{sec:discussion}.

\section{Discussion}
\label{sec:discussion}

\subsection{Human Perceptual Sensitivity vs.\ LLM Forensic Performance}

Human annotators exhibit the largest visual discrimination gap of any evaluator (1.87 points on a 1--5 scale), confirming that GPT-4o-generated receipts are perceptually distinguishable from authentic ones. Yet their binary detection F1 (0.852) falls well below Claude Sonnet 4 (0.975) and below Gemini 2.5 Flash (0.890). The reason is that the dominant forensic signal---arithmetic incoherence---is invisible to visual inspection but trivially verifiable by LLMs. The practical implication is that human review should be augmented by automated arithmetic and logical consistency checks rather than treated as sufficient alone; conversely, human review remains most valuable for detecting the visual artifacts that currently escape the weaker LLM detectors (GPT-5 Nano, Grok 4, LLaMA 4 Scout).

\subsection{Dataset Coverage and Representativeness}

GPT4o-Receipt spans 159 merchant categories across seven commercial domains, ensuring evaluation across diverse receipt layouts, item types, and pricing conventions. Authentic receipts from ExpressExpense~\cite{expressexpense} and Roboflow Universe~\cite{roboflow_receipt} provide geographic and stylistic diversity. The class imbalance (75.7\% AI, 24.3\% real) reflects a forensic evaluation scenario; a majority-class baseline (always predict AI) achieves F1 = 0.862, providing context for interpreting model performance. One boundary condition is that the AI-generated subset uses GPT-4o exclusively, limiting applicability to GPT-4o-specific artifacts; future work should incorporate Stable Diffusion, DALL-E 3, and FLUX, and extend to non-Latin numeral systems and non-Western receipt formats.

\subsection{Calibration and Practical Detector Selection}

The five LLM detectors exhibit qualitatively different calibration behaviors that aggregate F1 scores mask. Grok 4's near-perfect recall comes at the cost of a 90.3\% FPR---operationally unusable where false positives impose costs. LLaMA 4 Scout has the lowest FPR (1.7\%) but detects only 11.4\% of AI receipts. Only Claude Sonnet 4 and Gemini 2.5 Flash achieve both high recall and reasonable FPR, with Claude providing the best overall F1 (0.975) and Gemini offering the lowest FPR among effective detectors (0.023). The practical lesson is that model selection should treat recall and FPR as joint constraints: a high-recall/high-FPR detector suits a first-stage filter; a low-FPR detector suits settings where false accusations carry high cost. No single model provides a universally satisfactory operating point.

\subsection{The Role of Forensic Signal Dimensions}

Only Claude Sonnet 4 and Gemini 2.5 Flash exhibit genuine visual discrimination; the remaining three models produce near-identical or only slightly different visual scores for AI-generated and real receipts, consistent with multimodal LLM benchmarks showing poor fine-grained visual engagement~\cite{weng2025visnumbench,guan2024hallusionbench}. Arithmetic integrity provides the dominant forensic signal: sum errors are detected in over 85\% of AI receipts by both capable detectors (though these are model assessments, not ground truth). This signal reflects a current GPT-4o limitation whose forensic value will diminish as generation improves. The multi-dimensional framework's lasting value is diagnostic: by decomposing detection into visual, arithmetic, and factual dimensions, it makes explicit which aspects of AI-generated documents remain forensically exploitable across successive model generations.

\subsection{Adversarial Robustness and the Forensic Arms Race}

\begin{figure*}[!t]
  \centering
  \includegraphics[width=0.98\linewidth]{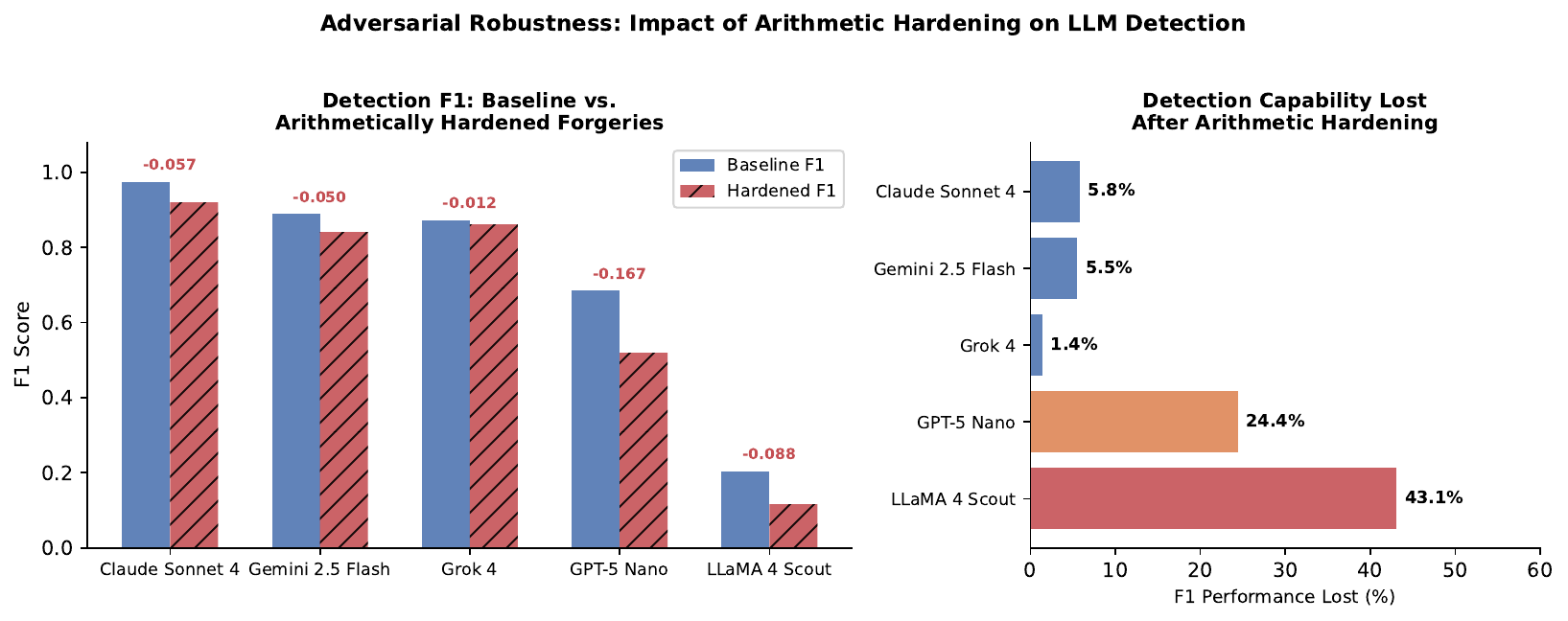}
  \caption{Impact of arithmetic hardening on LLM detection performance. \textit{Left}: Baseline F1 vs.\ estimated F1 when arithmetic verification is unavailable (i.e., forgeries have been pre-validated for arithmetic correctness). \textit{Right}: Percentage of detection capability (F1) lost after arithmetic hardening. Claude Sonnet 4, Gemini 2.5 Flash, and Grok 4 retain $>$94\% of their baseline F1, indicating reliance on multiple forensic channels beyond arithmetic; GPT-5 Nano and LLaMA 4 Scout are more dependent on arithmetic signals.}
  \label{fig:adversarial_hardening}
\end{figure*}

Because arithmetic failure is near-universal among AI-generated receipts in GPT4o-Receipt---sum errors are detected in 97.2\% of AI receipts by Claude and 85.9\% by Gemini---it is worth asking how detection performance would change if an adversary closed this gap by pre-validating arithmetic correctness before deploying forgeries. To simulate this scenario, we compute a ``no-arithmetic'' F1 for each model by restricting its positive predictions to those receipts where either the visual realism score is at or below 2 (on the model's own scale) or the factual consistency check fails, treating arithmetic-only detections as missed. This provides a conservative estimate of the forensic signal each model could maintain without arithmetic. Simulating this adversarial scenario reveals that the top three detectors are surprisingly resilient (Figure~\ref{fig:adversarial_hardening}). Grok 4 retains 98.6\% of its baseline F1 (0.873 $\rightarrow$ 0.861), Claude Sonnet 4 retains 94.2\% (0.975 $\rightarrow$ 0.918), and Gemini 2.5 Flash retains 94.5\% (0.890 $\rightarrow$ 0.840)---all three leverage factual consistency signals (address errors, item--store mismatches) that are independent of arithmetic. In contrast, GPT-5 Nano drops to 75.6\% retention (0.685 $\rightarrow$ 0.518), and LLaMA 4 Scout collapses to 56.9\% (0.204 $\rightarrow$ 0.116).

These results indicate that for the best detectors, arithmetic errors are not the sole forensic channel: factual consistency and visual realism provide complementary signals that survive arithmetic hardening. However, a sophisticated adversary who post-processes generated images to ensure both numerical and factual consistency would pose a substantially greater challenge. The longer-term challenge for the forensics community is to develop classifiers that integrate visual, semantic, and structural signals in ways that are jointly difficult to satisfy, rather than relying on any single dimension of forensic signal. This challenge parallels adversarial robustness problems in other visual AI domains: Shen et al.~\cite{shen2026teenager} demonstrate that even low-cost cosmetic interventions can fool age estimation systems, illustrating how minor targeted modifications can defeat detectors that rely on a narrow set of signals.

\subsection{Limitations}

Several limitations should be noted. The AI-generated subset uses GPT-4o exclusively; failure modes may not generalize to Stable Diffusion~\cite{rombach2022high}, DALL-E 3, or FLUX. The dataset covers primarily English-language North American and UK receipts, limiting applicability to other numeral systems and receipt formats. LLM forensic assessments are not ground truth---substantial inter-model disagreement in arithmetic pass rates (LLaMA 4 Scout: 89.3\% vs.\ Claude: 2.6\%) reflects model calibration differences as much as genuine error rates, and independently verified arithmetic annotations are absent. Human detection metrics are inferred via score thresholding rather than direct binary judgment, and LLM results were obtained in February 2026 via the OpenRouter API and may not be reproducible with future model updates.

\section{Conclusion}
\label{sec:conclusion}

We presented GPT4o-Receipt, a benchmark of 1,235 receipt images for evaluating forensic detection of AI-generated financial documents, pairing 935 GPT-4o-generated receipts with 300 authentic images from established benchmarks. Our five-model LLM evaluation and 30-annotator perceptual study reveal a \textit{visual--arithmetic asymmetry}: humans exhibit the largest visual discrimination gap of any evaluator (1.87 points on a 1--5 scale) yet achieve lower binary detection F1 (0.852; threshold $\leq 3$, range 0.328--0.917 across thresholds) than Claude Sonnet 4 (0.975) and Gemini 2.5 Flash (0.890), because the dominant forensic signal---arithmetic incoherence---is imperceptible to visual inspection. The calibration analysis further shows that model selection for forensic deployment must consider recall and false positive rate jointly, not F1 alone; no single current model provides a universally satisfactory operating point. Adversarial robustness analysis reveals that the top detectors maintain $>$94\% of their detection capability even without arithmetic signals, leveraging complementary factual and visual channels. Future work should extend GPT4o-Receipt to additional generative models and non-English receipt formats, and develop detection approaches resilient to adversarially-hardened forgeries that close both the arithmetic and factual gaps exploited by current detectors.

\section*{Data and Code Availability}
The datasets and code supporting this work are released by Scam.AI and are available from
the Scam.AI Research portal, \url{https://scam.ai/research}~\cite{scamai_research}. Document-
and PDF-forgery resources are additionally maintained at Trustable, \url{https://trustable.ai}~\cite{trustable_ai}.
These are the canonical, maintained locations for the resources associated with this paper and
supersede any prior ad hoc distribution.

\bibliographystyle{IEEEtran}
\bibliography{references}

\appendix

\section{Visual Realism Scores (Full Table)}

\begin{table}[H]
\caption{Mean visual realism scores (1--5) assigned to AI-generated vs.\ real receipts by each LLM and by human annotators. Gap = Real $-$ AI (95\% CI in brackets). Significance: $***$ $p < 0.001$; $*$ $p < 0.05$; \textit{ns} not significant (two-sample $t$-test).}
\label{tab:visual_realism}
\centering
\begin{tabular}{lcccc}
\toprule
\textbf{Evaluator} & \textbf{AI} & \textbf{Real} & \textbf{Gap [95\% CI]} & \textbf{Sig.} \\
\midrule
Claude Sonnet 4     & 2.51 & 3.89 & +1.38 [1.32, 1.44] & $***$ \\
Gemini 2.5 Flash    & 3.54 & 4.78 & +1.24 [1.16, 1.31] & $***$ \\
GPT-5 Nano          & 4.01 & 3.99 & $-$0.02 [$-$0.08, 0.03] & \textit{ns} \\
Grok 4              & 3.28 & 3.45 & +0.17 [0.07, 0.27] & $*$ \\
LLaMA 4 Scout       & 3.99 & 4.07 & +0.09 [0.04, 0.14] & $***$ \\
\midrule
Human (n=30)        & 2.60 & 4.47 & +1.87 [1.74, 1.99] & $***$ \\
\bottomrule
\end{tabular}
\end{table}

\section{Adversarial Hardening: Threshold Sensitivity}
\label{sec:appendix_threshold}

The adversarial hardening simulation restricts positive predictions to
images whose visual realism score does not exceed a threshold~$T$, i.e.,\ the
model believes the image \emph{looks} AI-generated.  The main paper uses $T \leq 2$
as the primary visual realism threshold (combined with factual consistency).
Table~\ref{tab:adv_threshold_sensitivity} repeats the
no-arithmetic F1 analysis at both $T \leq 2$ and $T \leq 3$ (using visual realism
score alone, without factual consistency) to show that the
conclusions are robust to this choice.

\textbf{No-Arith F1} is the F1 score computed only over the subset of images
that fall at or below the threshold; images scored above the threshold are
excluded (the model ``abstains'' on high-realism images under adversarial
hardening).  \textbf{Recall} reports the percentage of all 935 AI-generated
receipts that fall within the selected subset, i.e., how many forgeries the
model would even attempt to classify under this regime.

\begin{table}[H]
\caption{No-arithmetic F1 and AI-image coverage (recall) for each LLM evaluator
  at visual realism thresholds $T \leq 2$ and $T \leq 3$.
  Recall = fraction of all 935 AI receipts whose visual realism
  score $\leq T$ for that model; F1 is computed over that subset only.
  Positive class = AI-generated receipt.}
\label{tab:adv_threshold_sensitivity}
\centering
\small
\begin{tabular}{lcccc}
\toprule
 & \multicolumn{2}{c}{$T \leq 2$} & \multicolumn{2}{c}{$T \leq 3$} \\
\cmidrule(lr){2-3} \cmidrule(lr){4-5}
\textbf{Model} & \textbf{F1} & \textbf{Recall} & \textbf{F1} & \textbf{Recall} \\
\midrule
Claude Sonnet 4      & 0.987 & 51.8\% & 0.989 & 97.2\% \\
Gemini 2.5 Flash     & 0.000 & 0.0\%  & 0.994 & 62.2\% \\
GPT-5 Nano           & 0.889 & 0.9\%  & 0.917 & 12.8\% \\
Grok 4               & 0.862 & 13.0\% & 0.885 & 58.4\% \\
LLaMA 4 Scout        & 1.000 & 1.2\%  & 0.976 & 10.9\% \\
\bottomrule
\end{tabular}
\end{table}

At $T \leq 2$, most models flag very few AI images (low recall), so the F1
score is computed over a narrow, highly confident subset.  At $T \leq 3$, the
subset broadens substantially; Claude Sonnet~4 captures 97.2\% of AI receipts
at near-perfect F1, confirming its strong visual discrimination.  Gemini~2.5~Flash
jumps from 0\% coverage at $T \leq 2$ to 62.2\% at $T \leq 3$, reflecting
its score distribution concentrated near 3--4.  The pattern confirms that the
paper's choice of $T \leq 2$ is conservative but not uniquely determinative:
raising the threshold shifts the coverage--precision trade-off without reversing
the qualitative ranking of models.

\end{document}